\documentclass[conference]{IEEEtran}
\IEEEoverridecommandlockouts
\usepackage{cite}
\usepackage{amsmath,amssymb,amsfonts}
\usepackage{algorithmic}
\usepackage{graphicx}
\usepackage{textcomp}
\usepackage{xcolor}
\usepackage{threeparttable}
\usepackage{dblfloatfix}
\usepackage{adjustbox}
\usepackage{caption}
\usepackage{subcaption}
\usepackage{multirow}
\def\BibTeX{{\rm B\kern-.05em{\sc i\kern-.025em b}\kern-.08em
    T\kern-.1667em\lower.7ex\hbox{E}\kern-.125emX}}
\usepackage{comment}
\usepackage{authblk}
\begin{document}

\title{Noise-Tolerance GPU-based Age Estimation Using ResNet-50}
\author[1]{Mahtab Taheri}
\author[2]{Mahdi Taheri}
\author[1]{Amirhossein Hadjahmadi}

\affil[1]{Vali-e-Asr University of Rafsanjan, Kerman, Iran}
\affil[2]{Tallinn University of Technology, Tallinn, Estonia}
\IEEEoverridecommandlockouts \IEEEpubid{\makebox[\columnwidth]{978-1-6654-7355-2/22/\$31.00~\copyright2022 IEEE \hfill} \hspace{\columnsep}\makebox[\columnwidth]{ }}
\maketitle
 \IEEEpubidadjcol
\begin{abstract}
The human face contains important and understandable information such as personal identity, gender, age, and ethnicity. In recent years, a person's age has been studied as one of the important features of the face. The age estimation system consists of a combination of two modules, the presentation of the face image and the extraction of age characteristics, and then the detection of the exact age or age group based on these characteristics. So far, various algorithms have been presented for age estimation, each of which has advantages and disadvantages. In this work, we implemented a deep residual neural network on the UTKFace data set. We validated our implementation by comparing it with the state-of-the-art implementations of different age estimation algorithms and the results show 28.3\% improvement in MAE as one of the critical error validation metrics compared to the recent works and also 71.39\% MAE improvements compared to the implemented AlexNet. In the end, we show that the performance degradation of our implemented network is lower than 1.5\% when injecting 15 dB noise to the input data (5 times more than the normal environmental noise) which justifies the noise
tolerance of our proposed method.         
\end{abstract}


\vspace{-0.2cm}\section{Introduction}
Nowadays, a lot of information can be obtained from face images, including identity, gender, age, race, and even emotional states \cite{han2014demographic}. Age estimation is one of the most important and sensitive applications that rely on the information extracted from face images. Automatic age estimation systems can prevent mental problems and personality disorders in teenagers in the future by limiting the access of people who are under the legal age to inappropriate sites or online purchase pages of inappropriate products \cite{mahjabin2019age}. Age estimation can also make pursuing the suspects easier in a way that it can make it possible to limit the gallery by using the estimated age filter on the data set, and suspects can be identified more accurately and quickly \cite{6920084}. Age estimation can also help to attract customers to stores, by implementing smart shopping baskets. This approach can offer popular products to particular customers of desired age category. Even though, age determination plays an important role in different security-critical applications, determining one’s actual age based on their appearance is a challenging task \cite{yi2015age}. 
Face can remain young due to various factors such as weather, stress, genetics, etc. \cite{6920084}. Therefore, even without any prior knowledge, one can notice that accurate age estimation is far more complicated than classifying different age groups.
Although this task is relatively more complicated, it is also more important for criminology and identity recognition. Therefore, handing over this task to machine learning algorithms to overcome this issue through their superior advancements,  becomes very interesting. 
Artificial Neural Networks (ANN), especially Deep Neural Networks, have made significant progress and become a hot topic in recent years. The basis of the work of an ANN is to imitate the human mind to learn different subjects. ANNs have three main layers \cite{da2017danilo}: (1) input layer: which is responsible for receiving data and features as the input data. (2) hidden layers: which are composed of neurons, are responsible for processing and analysis. (3) output layer: also consists of neurons and produces the output of the entire network, which is based on the processes of the previous layer. In general, several different network topologies and architectures are proposed so far (e.g., Convolutional Neural Networks (CNNs). These networks extract the features of more complex shapes. CNNs have different architectures including (1) AlexNet (2) VGG (3) ZfNet (4) GoogleNet (5) ResNet \cite{alzubaidi2021review}. Some works try to improve the performance of CNNs by changing networks' characteristics and architecture, such as changing the depth of the network, changing the cost function, etc. Our contribution in this paper is the adoption and implementation of ResNet-50 architecture for age estimation to tackle the Mean Absolute Error (MAE) rate as our main evaluation metric. To the best of our knowledge, it is the first time that ResNet-50 network is being used for this application. We also implemented AlexNet as our base comparison network and show our superior results compared to other state-of-the-art works as well.

The rest of the paper is organized as follows. An analysis of
Related Works in Section II is followed by a background on Residual Neural Networks presented in Section III. Our age estimation methodology is presented in section IV. The results, along
with their discussion, are presented in section V. Finally, this
work is concluded in Section VI.

\vspace{-0.2cm}\section{Related Work}
The importance of applying different methods for age estimation is advocated in different works.
Kang et al. gained a good performance against various motion blur effects by implementing ResNet-152. Due to the fact that Resnet-152 does not need pre-classification, it simplifies the system and reduces the error rates on PAL and MORPH data sets \cite{kang2018age}.
Arna Fariza et al. presented their work to increase accuracy by implementing their proposed network, ResNeXt-50, on the UTKFace data set\cite{fariza2019age}.
Mohammed Kamel Benkaddour et al. by changing the structure and depth of networks, defined 3 CNN models with different filter sizes and used them for age and gender estimation \cite{benkaddour2021human}.
Xiaoqiang Li et al. presented a small model-size network called Group-aware Contrastive Network(GACN). This network, which was tested on Morph II and FG-Net data sets, shows acceptable stability against environmental changes and also achieved competitive performance in public data sets\cite{li2022robust}.
Hao Liu et al. improved the age estimation accuracy by presenting the similarity-aware deep adversarial learning (SADAL) on the Morph and FG-Net data sets, by using the similarity-aware membership, which is based on the three criteria of hard negative mining, the reconstruction loss, and generating fake sample, and the variational deep adversarial learning (VDAL) framework. The VDAL framework aims to factorize each face sample to the intra-class variance distribution and invariant class center\cite{liu2020similarity}. 
Although there are many works trying to address age estimation, we provided a comparison between the work presented in this paper, the AlexNet network that is implemented by authors as the base comparison network, and also the state-of-the-art works i.e., ResNeXt-50\cite{fariza2019age}, Linear Regression \cite{fariza2019age}, WAS \cite{cootes2001active}, AAS\cite{cootes2001active}, AGES\cite{cootes2001active}, KNN\cite{lanitis2002toward}, BP\cite{lanitis2002toward}, SVM \cite{lanitis2002toward}, CMT learning\cite{yoo2018deep}. 

\vspace{-0.2cm}\section{Background}

\begin{figure}[h]
    \centering
\includegraphics[width=0.75\columnwidth]{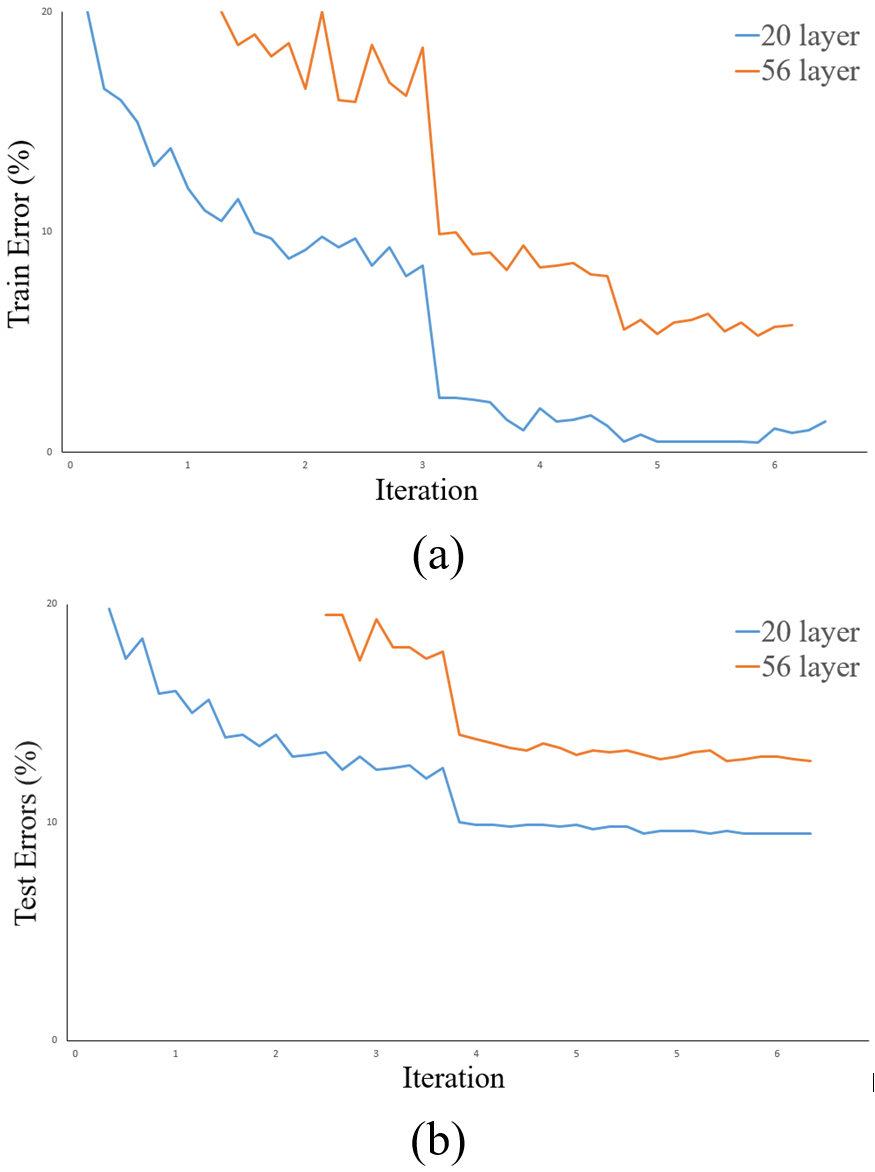}
         \caption{Impact of increasing the layers of a network on (a) training error (b) test error}
         \label{trtif} \label{depth1}
     \end{figure}

Some issues like vanishing gradient and higher training errors make it difficult to train very deep neural networks \cite{punyani2020neural}. Fig. \ref{depth1} is illustrating that in many cases, increasing the layers of a network does not help with improving the performance of the network. ResNet architecture, on the other hand, by using skip connections allows it to increase the depth of the network without facing the mentioned issues.
\vspace{-0.2cm}\subsection{ResNet-50}
\begin{figure}[h]
    \centering
\includegraphics[width=0.5\columnwidth]{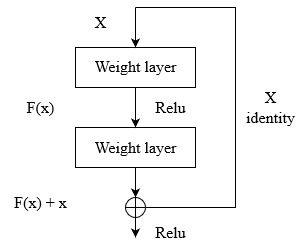}
    \caption{Residual Block \cite{he2016deep}}
    \label{fig:r}
\end{figure}

Deep Residual Networks \cite{he2016deep} are an innovation in recent years that allow the network to grow its layers up to thousands of layers, without causing any performance degradation, and help with solving training issues like gradient-vanishing because of their architecture.
As shown in Figure \ref{fig:r}, shortcut connections or redundant connections are the solutions provided by the ResNet architecture to solve the problem of deep networks, which enables the network to have numerous layers. Residual neural networks have shortcut connections that pass data through one or more layers and provide the capability to bypass the rest.; In fact, they take shortcuts and connect a layer to a further layer. Therefore, the residual block allows the gradient to flow unimpeded through the shortcut connection to the previous layer. This residual block explicitly causes lower training errors.
\vspace{-0.2cm}\subsection{AlexNet}
This architecture includes five convolution layers followed by max pooling and 3 fully connected layers. This architecture also benefits Relu as the activation function and uses the Dropout layer to avoid overfitting.

\vspace{-0.2cm}\section{Age Estimation Methodology}

ResNet contains two main types of blocks whose use depends on the differences and similarities in input/output dimensions. 
The identity block is a standard block that corresponds to a state where the input activation has the same dimensions as the output activation. If the input and output dimensions are not the same, the convolutional block is used. The difference between this block and the identity block is that there is a CONV2D layer in the shortcut path. 
These two blocks are implemented separately and then used in the implemented ResNet-50 network at this work. In general, this architecture has five hidden layers, each layer has a convolutional block, a number of identity blocks, and a max pooling only in the first layer. In this paper, the linear activation function and adam optimizer are used.
\begin{figure*}[h!]
     \centering
     \begin{minipage}[b]{0.4\textwidth}
         \centering
         \includegraphics[width=1\textwidth]{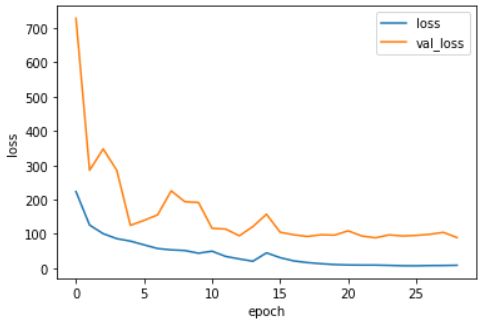}
         \caption{Proposed method: Loss and Val-Loss results}
         \label{res1}
     \end{minipage}
     \hfill
     \begin{minipage}[b]{0.4\textwidth}
         \centering
         \includegraphics[width=1\textwidth]{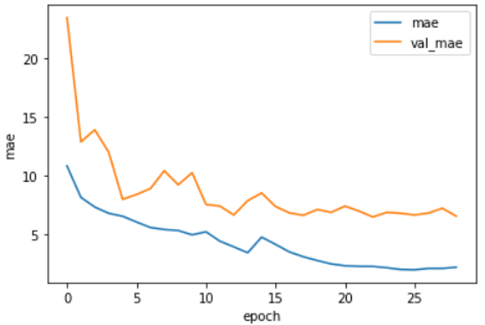}
         \caption{Proposed method:  MAE and Val-MAE results}
         \label{res2}
     \end{minipage}
     \centering
     \begin{minipage}[b]{0.4\textwidth}
         \centering
         \includegraphics[width=1\textwidth]{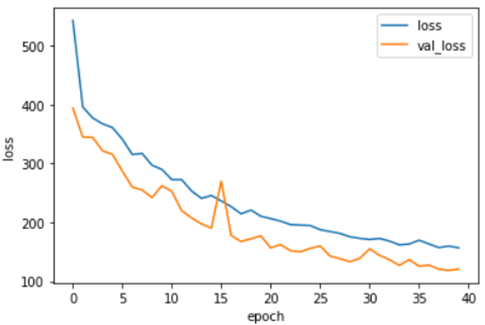}
       \caption{AlexNet: Loss and Val-Loss results}
         \label{res3}
     \end{minipage}
     \hfill
     \begin{minipage}[b]{0.4\textwidth}
         \centering
         \includegraphics[width=1\textwidth]{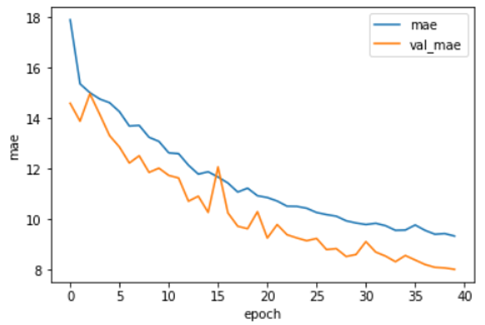}
        \caption{AlexNet: Loss and Val-Loss results}
         \label{res4}
     \end{minipage}
\end{figure*}

\vspace{-0.2cm}\section{Results}
In this section, we provide a detailed setup, results, and discussion of our proposed work.

\vspace{-0.2cm}\subsection{Experimental Setup}
All networks in this work are implemented, trained, and tested in python version 3.6.8 using keras library v2. 
All the simulations are performed on the Nvidia GeForce RTX 3080 GPU.

\vspace{-0.2cm}\subsection{Data set}
UTKFace is a collection of faces from different nationalities, genders, and ages. This data set is a set of large-scale face images that covers both genders, men and women, with different nationalities, Asian, Indian, Hispanic, white, and black, at different ages (0-116). It also provides a cropped version of the original images for better feature recognition. Different challenges of using this data set like uneven age distribution, image brightness, and different image positions make it an excellent benchmark to evaluate the performance of the age estimation algorithm.

Needless to say, pre-processing plays an essential role in data processing tasks and helps to improve the results significantly. There are various methods for this sake, including removing outliers and non-existent data, data augmentation, and normalization. For this work, images are normalized and 16593 images are randomly selected as training data and 7112 are grouped for test data, out of 23705 images in the data set. 


\vspace{-0.2cm}\subsection{Experimental Results}
To obtain the results, AlexNet is implemented along with ResNet-50 as our comparison base due to its efficiency and good performance. 
\begin{table}[h!]
\centering
\caption{ResNet-50 vs. AlexNet train and test efficiency comparison}
\label{tab:propose result}
\begin{tabular}{|c|c|c|c|c|}
\hline
Methods &
Loss &
MAE &
Val-Loss &
Val-MAE
  \\ \hline
AlexNet &
  160.17\ &
  9.43\ &
  119.45\ &
  7.96\
 \\ \hline
Propose method &
  8.58\ &
  2.18\ &
  89.41\ &
  6.53\
\\ \hline
\end{tabular}
\end{table}
\begin{figure}[h!]
\centering\includegraphics[width=0.5\textwidth]{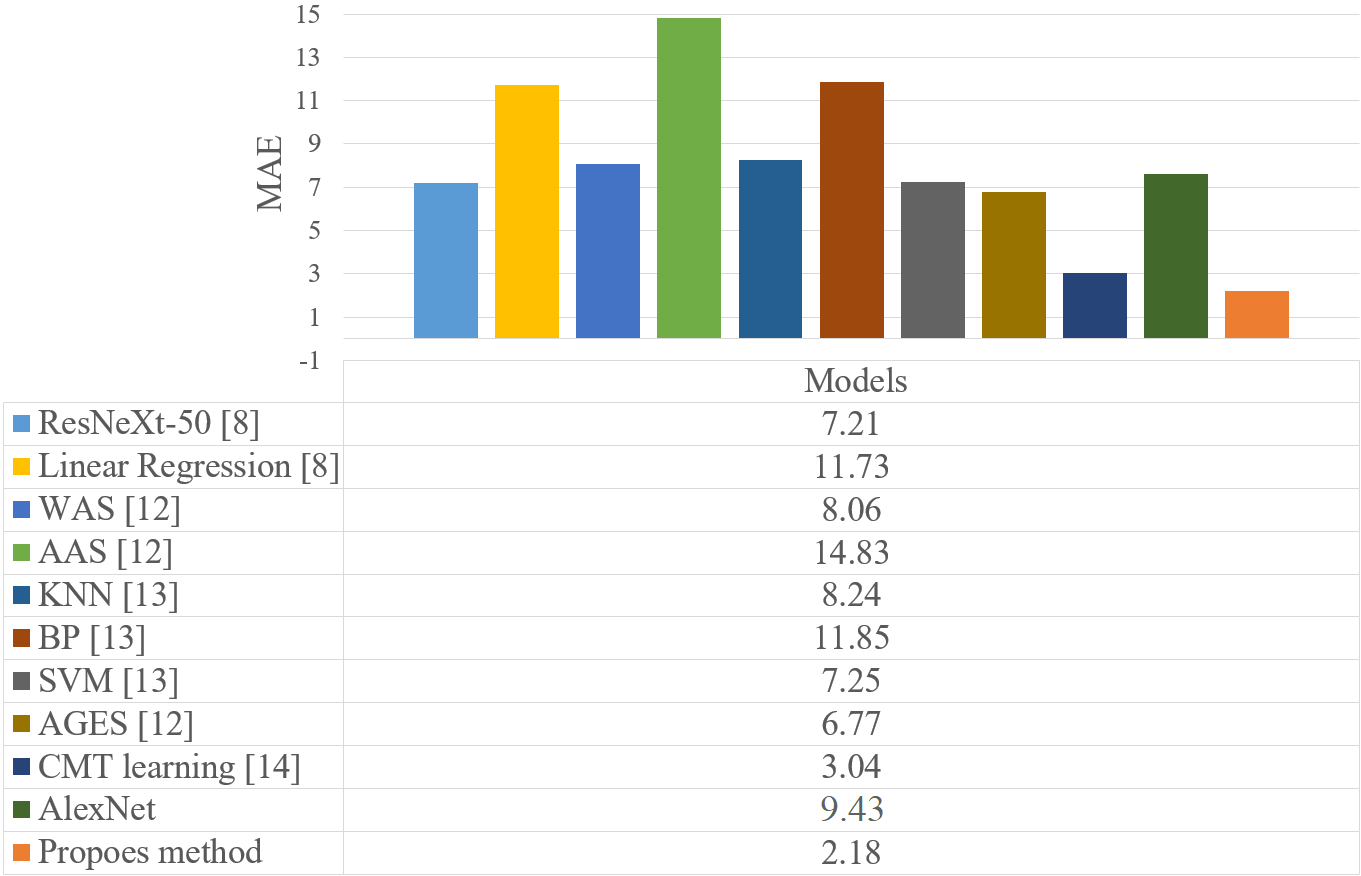}
    \caption{Training MAE comparison between different models}
    \label{fig:t.}
\end{figure}
Both networks are implemented on the same data set and are performed on the same hardware for a fair comparison.
Results reported in Fig. \ref{res1} - \ref{res4} demonstrate Loss, Val-Loss, MAE, and Val-MAE reports for ResNet-50 and AlexNet during the training process. In Table. \ref{tab:propose result}, different training and test error metrics are provided to compare AlexNet and ResNet-50. As it is shown in this table, the proposed method is showing ~71.39\% improvement in MAE, and a significant improvement in the other metrics compared to AlexNet.

The superiority of the proposed method compared to the state-of-the-art works is also shown in Fig. \ref{fig:t.}. This figure shows a noticeable improvement in the training performance of our implemented ResNet-50 network compare to the other network architectures. 

\vspace{-0.2cm}\subsection{Noise simulation}

In the meantime, when our network is being performed on a GPU, environmental noise is affecting the data storage and might impact our network inference performance. Such noise stems from the environment's white noise, electrical noise, etc. Hence, age estimation in critical applications should be noise tolerant, so their performance does not degrade in real-world scenarios. Worst-case initial SNR is about 2 dB in such real-life environments \cite{shekar2019wavelet}; however, we have simulated a more extensive range of probable noise values to see if our network is tolerant against these values.
various powers of white noise in our model are used to create the noisy data set: ranging from 2 dB to 15 dB. Then, the inference setup is used, and the performance degradation is calculated for various faulty input data of face images.
\begin{figure}[h]
    \centering
    \includegraphics[width=1\columnwidth]{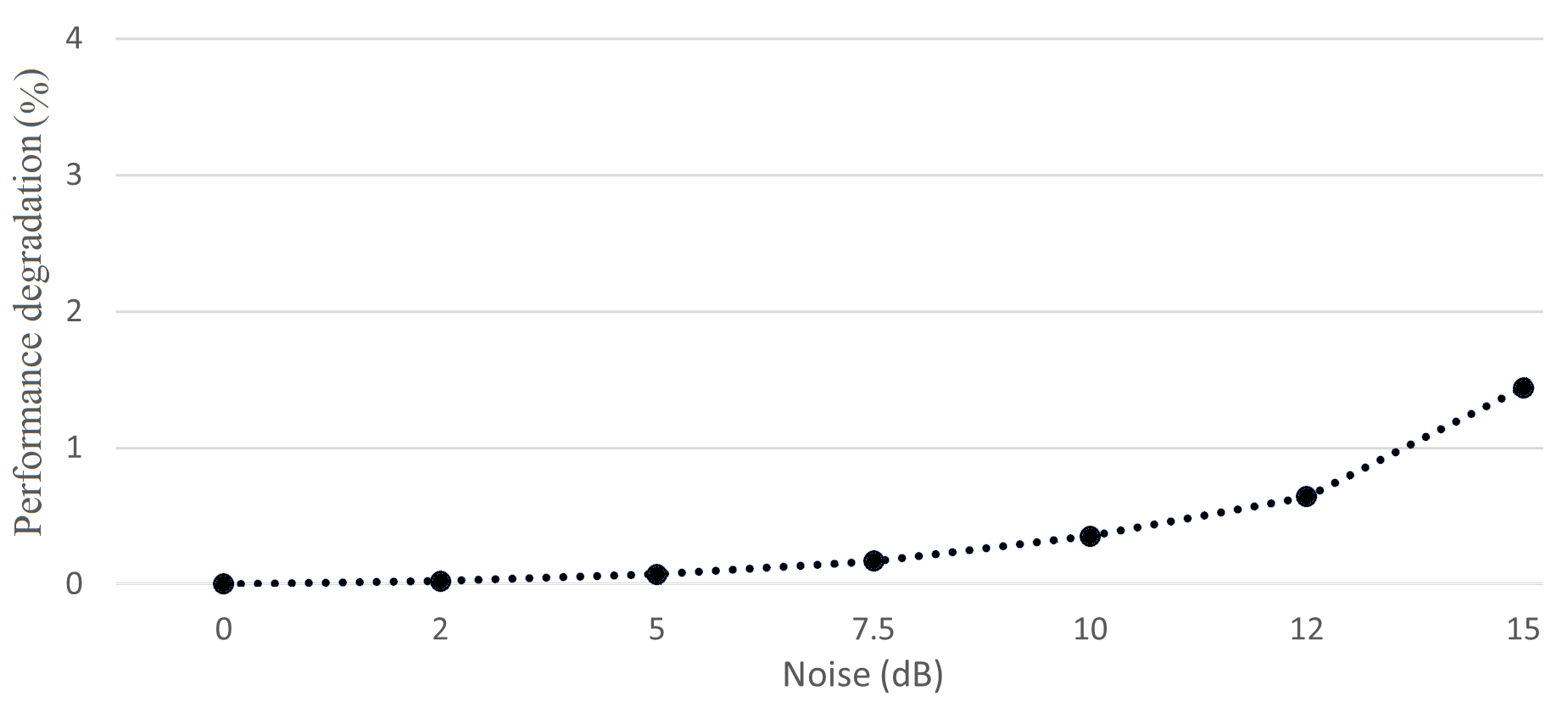}
    \caption{Percentage of performance degradation per noise injected in dB}
    \label{fig:noise_red}
\end{figure}

Finally, average noise reduction is calculated throughout our test data and the results are reported in Fig. \ref{fig:noise_red}. As demonstrated in this figure, the performance degradation is lower than 1.5\% when injecting 15 dB noise to input data (5 times the normal environmental noise); we observe 0.02\% and 0.07\% performance degradation for 2dB and 5dB noise, respectively. These results justify the noise tolerance of our proposed method

\vspace{-0.2cm}\section{Conclusion}
 In this work, we implemented a GPU-based deep residual neural network on the UTKFace data set. We validated our implementation by comparing it with different state-of-the-art implementations of age estimation
methods. Provided results show a significant improvement in network training results. Based on the various training error metrics reported in this paper, it is shown that our proposed method can obtain about 28.3\% improvement in MAE as one of the critical error validation metrics compared to the recent works and also 71.39\% MAE improvements compared to the implemented AlexNet. We also show that the performance degradation of our implemented network is lower than 1.5\% when injecting 15 dB noise to the input data (5 times more than the normal environmental noise) which justifies the noise
tolerance of our proposed method. 


\bibliographystyle{IEEEtran}


\bibliography{refrence}

\end{document}